\documentclass[10pt,twocolumn]{article}

\usepackage[margin=1in,columnsep=0.25in]{geometry}
\usepackage[T1]{fontenc}
\usepackage[utf8]{inputenc}
\usepackage{microtype}
\usepackage{lmodern}

\usepackage{amsmath,amssymb,amsthm}

\usepackage{booktabs}
\usepackage{tabularx}
\usepackage{multirow}

\usepackage[colorlinks=true,linkcolor=blue!70!black,citecolor=blue!70!black,urlcolor=blue!70!black]{hyperref}

\usepackage{titlesec}
\titleformat{\section}{\normalfont\large\bfseries}{\thesection.}{0.5em}{}
\titleformat{\subsection}{\normalfont\normalsize\bfseries}{\thesubsection.}{0.5em}{}
\titleformat{\subsubsection}{\normalfont\normalsize\itshape}{\thesubsubsection.}{0.5em}{}

\usepackage{abstract}

\usepackage{caption}
\usepackage{graphicx}
\usepackage{xcolor}
\usepackage{enumitem}
\setlist{noitemsep, topsep=2pt}

\usepackage[numbers,sort&compress]{natbib}
\title{\textbf{Latent Communication Between Language Model Agents:\\
       Channels, Alignment, and the Limits of Text}}

\author{%
  [Markus Wenzel]\\
  \small [Constructor University]\\
  \small \texttt{[mwenzel@constructor.university]}
}

\date{}

\begin{document}

\twocolumn[
  \maketitle
  \begin{@twocolumnfalse}
    \begin{abstract}
Multi-agent systems (MAS) are utilized in many contexts ranging from private personal assistants over supportive tools in many professions to independent or human-steered research systems. Those MAS rely on inter-agent communication, usually implemented by clear-text message passing. We hypothesize that Large Language Models may have a world model at their disposal that exceeds expressibility in text when complex concepts need to be communicated. Our aim is to approach a proof of this hypothesis with structured experiments.

In this work, we show that LLM agents communicating via text lose information, which we quantify via Sparse Autoencoder (SAE) feature analysis. We construct three communication channels ("dense latent" directly injecting the LLM's internal representation, "SAE-sparse" transferring a sparse code, and "text" being the usual prompt) and measure concept-discriminating information in each. We first show that the SAE-sparse channel retains a 99.4\% probe accuracy at 28-fold compression over the dense-latent channel vs 80.4\% for the text channel.

We then proceed to examine the same for cross-architecture communication by using sparse latent space alignment. We find even for a simple Procrustes alignment a 92\% top-1 retrieval between Llama and Mistral. Using a text round-trip (serialize and re-encode), we perform feature survival analysis to find that text serialization destroys 88\% of SAE features, replacing them with a different feature set. We attribute the loss to \textit{identity replacement}, not attenuation, but also find that the lost features do not carry the bulk of semantic information. By our analysis, we were able to attribute a 3--10\,pp performance penalty to the linear Procrustes alignment. These numbers improve with nonlinear alignment methods.

We lastly progress to a task-level evaluation, finding that the latent channel matches the text channel on cross-lingual concept tasks but never exceeds it. In our framework, text augmentation with latent features provides no benefit, leading us to negative conclusions for the initial hypothesis: lost features mostly or completely encode surface form, not task-relevant semantics. However, representational convergence between architectures is real and potentially exploitable. To pinpoint the practical advantage of latent communication over a text channel, deeper tasks eliciting complex concepts and an corresponding analysis framework are needed which are both beyond our current approach.
    \end{abstract}
    \vspace{1em}
  \end{@twocolumnfalse}
]

% ─────────────────────────────────────────────────────────────────────────────
\section{Introduction}
\label{sec:intro}

Multi-Agent Systems utilizing Large Language Models (LLMS) as atomic Agents have become a central paradigm in applied AI, with frameworks such as AutoGen and CrewAI enabling complex pipelines in which multiple agents collaborate to solve tasks. In virtually all current implementations, agents communicate by exchanging natural language messages or structured information expressed in JSON etc. In this communication paradigm, agents generate and read messages transported in plain text. This bears advantages: text is human-readable, debuggable, and compatible with any model that understands language.

We hypothesize that text communication might be lossy in a way limiting inter-agent communication. When an agent processes a complex input and forms a high-dimensional internal representation of a concept, it must compress that representation into a sequence of tokens before another agent can receive it. The receiving agent then reconstructs its own internal representation from those
tokens. Nothing guarantees that the two representations are equivalent, or even similar. Our contribution seeks to quantify how much is lost, whether the loss matters, and whether a better alternative exists.

A compelling recent hypothesis suggests that the loss may be structurally
bounded. The Platonic Representation Hypothesis~\cite{huh2024platonic} proposes
that diverse neural networks trained on sufficient data converge toward
compatible representational geometries — a shared latent world model that
emerges independently of architecture or training objective. If this is true,
two language models will develop activation spaces that are geometrically
alignable, potentially enabling direct latent-space communication that bypasses
the text bottleneck entirely.

This paper pursues the testing of alternative communication channel systematically. We construct and compare
three communication channels — "dense latent" (the full activation vector),
"SAE-sparse" (a sparse bottleneck via pre-trained Sparse Autoencoders, SAEs), and text
(the standard text-based concept serialization baseline) — and measure how much
concept-discriminating information each preserves. 
We characterize information
loss at the SAE feature level (associated with Claim 2 below -- C2), 
measure channel fidelity via linear probes (C1), 
and test cross-architecture latent communication via Procrustes alignment (C3). 
Finally, we evaluate all channels on a battery of concept-identification
and discrimination tasks, including cross-lingual variants designed to favour
the architecture-agnostic structure of the latent channel.

% TODO: Should this part even be here already?
% TODO: CCA not mentioned!
Our findings are nuanced. Representational convergence between Llama 3.1
8B-Instruct and Mistral 7B is unambiguously confirmed: Procrustes-aligned dense
vectors achieve 92\% top-1 concept retrieval at 140 anchor concepts, against a
0.87\% random baseline. SAE feature analysis reveals that text serialization
destroys 88\% of original SAE features — but augmentation experiments establish
that these lost features encode surface form rather than task-relevant
semantics. On every tested task type, the text channel matches or outperforms
the latent channel by 3--10\,pp. We present this as an honest negative: the
infrastructure for latent communication is sound, but a practical advantage over
text requires tasks beyond current textual expressibility.

Our experiments follow a gated sequence designed to isolate the contribution of
each component. Two prerequisite \textit{gates} must pass before the main
claims are tested: semantic concepts must be linearly separable in the sender's
activation space, and the SAE feature space must faithfully capture semantic
relatedness (\S\ref{sec:baselines}). We then evaluate three claims: \textbf{C1}, that the
SAE-sparse channel preserves concept-discriminating information lost by text
(method description in \S\ref{sec:probes}, results in \S\ref{sec:c1}); \textbf{C2}, characterising what exactly text serialisation
destroys at the feature level (\S\ref{sec:survival-method}, \S\ref{sec:c2}); and \textbf{C3}, that
cross-architecture alignment enables latent communication between separately
trained models (\S\ref{sec:alignment-method}, \S\ref{sec:c3}). Task-level evaluation tests whether these
representation-level differences translate to downstream performance (\S\ref{sec:task-protocol},
\S\ref{sec:task-results}), and text augmentation resolves the consequential question: do the
features that text destroys carry suppressed semantic content (\S\ref{sec:augmentation-method}, \S\ref{sec:augmentation})?

\section*{Related work and own contributions}

\paragraph{Representational convergence.}
The Platonic Representation Hypothesis~\cite{huh2024platonic} proposes that
neural networks with sufficient capacity converge toward a shared kernel
estimating the statistical structure of the world. Cherti et
al.~\cite{cherti2023clip} provide complementary evidence from vision-language
models: CLIP models trained at increasing scale show monotonically improving
agreement between independently trained runs, consistent with convergence toward
a shared representation. Within a single architecture, Gromov et
al.~\cite{gromov2024layers} find that deeper layers in large transformers can be
pruned with minimal performance degradation, implying that representational
geometry stabilises well before the final layer. Despite these convergence
trends, the direct implications for cross-architecture communication have not
been tested quantitatively. Our cross-architecture alignment experiment
measures convergence in a retrieval accuracy benchmark. If our two models
indeed share representational structure (share a world model), a rotation of the embedding space fitted on a small anchor set should generalise to unseen concepts with far above-chance retrieval. This provides a hard threshold against which convergence can be evaluated, modeled as an initial gate in our experiments.

\paragraph{Model stitching and linear representation transfer.}
Lenc \& Vedaldi~\cite{lenc2015stitching} and Bansal et
al.~\cite{bansal2021similarity} demonstrated that internal representations of
deep networks can be aligned by learned linear maps, enabling functional
transfer across separately trained models. Orthogonal Procrustes was first
applied to word embeddings by Smith et al.~\cite{smith2017bilingual}, who showed
that bilingual dictionaries can be induced by aligning monolingual embedding
spaces with an orthogonal map and inverted-softmax retrieval. Conneau \&
Lample~\cite{conneau2018word} removed the need for seed dictionaries entirely,
using adversarial training followed by Procrustes refinement to achieve
unsupervised cross-lingual word embedding alignment. Our setting differs from
both model stitching and embedding alignment in two respects: there is no shared
task loss between sender and receiver, and our goal is lossless semantic
communication rather than functional substitution or word translation. The
Procrustes alignment we use is rotation-only — a strictly orthogonal map —
which tests a stronger form of geometric compatibility than general linear
stitching, extending the embedding-level work of Smith et al.\ and Conneau
\& Lample to contextual hidden states of LLMs.

\paragraph{Mechanistic interpretability and Sparse Autoencoders.}
The theoretical motivation for sparse factorisation comes from superposition:
Elhage et al.~\cite{elhage2022superposition} demonstrated that neural networks
can represent more features than they have dimensions by encoding them in
almost-orthogonal directions, governed by feature sparsity and importance. This
means that a 4,096-dimensional residual stream may encode tens of thousands of
features simultaneously — explaining why SAE dictionaries must be substantially
overcomplete (131K features for 4K-dimensional activations in our setup).
Bricken et al.~\cite{bricken2023monosemanticity} and Cunningham et
al.~\cite{cunningham2023sparse} showed that pre-trained SAEs factorise these
superposed representations into near-monosemantic features, primarily as a tool
for circuit-level reverse engineering of model computation. Templeton et
al.~\cite{templeton2024scaling} scaled SAEs to production models (Claude 3
Sonnet) and found that the resulting features are abstract, multilingual, and
multimodal — suggesting that SAE features capture semantic content rather than
surface form, which is the assumption underlying our feature survival analysis.
We repurpose SAEs here as a measurement instrument: comparing the sparse feature
sets elicited by two encoding paths for the same concept characterises
text-channel information loss at feature resolution — distinguishing attenuation
from replacement in a way that dense-space cosine similarity alone cannot
provide.

\paragraph{Cross-lingual representation alignment.}
Conneau et al.~\cite{conneau2020crosslingual} established that multilingual
pretraining produces cross-lingually aligned representations — the same concept
in different languages occupies compatible regions of a shared embedding space.
Pires et al.~\cite{pires2019multilingual} showed that this alignment emerges
even in Multilingual BERT without an explicit alignment objective, attributing
it to shared wordpieces and parameter sharing — a further instance of convergent
geometry arising from overlapping training statistics. Our cross-architecture
alignment is structurally analogous but operates along a different axis: where
cross-lingual alignment tests whether a single model represents the same concept
consistently across languages, we test whether two different architectures do so
in geometrically compatible ways. Both reflect invariances induced by shared
training statistics; the architectural case is the stronger test because it
imposes no joint training objective whatsoever.

\paragraph{Emergent communication and latent injection.}
Kottur et al.~\cite{kottur2017natural} showed that agents in referential games
develop effective but non-compositional communication protocols; natural language
does not emerge without explicit structural constraints. This motivates
exploring alternatives to emergent signalling: rather than learn communication
from scratch, our channels leverage the shared geometry that emerges from
independent pretraining. On the mechanistic side, Lester et
al.~\cite{lester2021prompt} demonstrated that frozen language models can be
steered by learned soft prompts — continuous vectors prepended to the input
embedding sequence. Our latent injection protocol is analogous: the receiver
model is frozen and the sender's activation vector is injected at a single layer
position, testing whether pre-existing representational compatibility can
substitute for the task-specific optimisation that prompt tuning requires.

% ─────────────────────────────────────────────────────────────────────────────
\section{Methods}
\label{sec:methods}

\subsection{Activation Extraction}
\label{sec:activation-extraction}

We use Llama 3.1 8B-Instruct as the sender model and Mistral 7B (base v0.1 and
instruction-tuned v0.3) as the receiver. Activations are extracted at
approximately 72--75\% of total depth: layer 23 of Llama and layer 24 of
Mistral, further referred to as the "extraction layer". This depth is motivated by prior work showing that intermediate-to-late
layers carry the highest density of semantic information. Activations are
extracted as the last-token hidden state of the residual stream in bf16
precision, yielding 4,096-dimensional vectors for both models.

\begin{figure*}[tb]
    \centering
    \includegraphics[width=1\columnwidth]{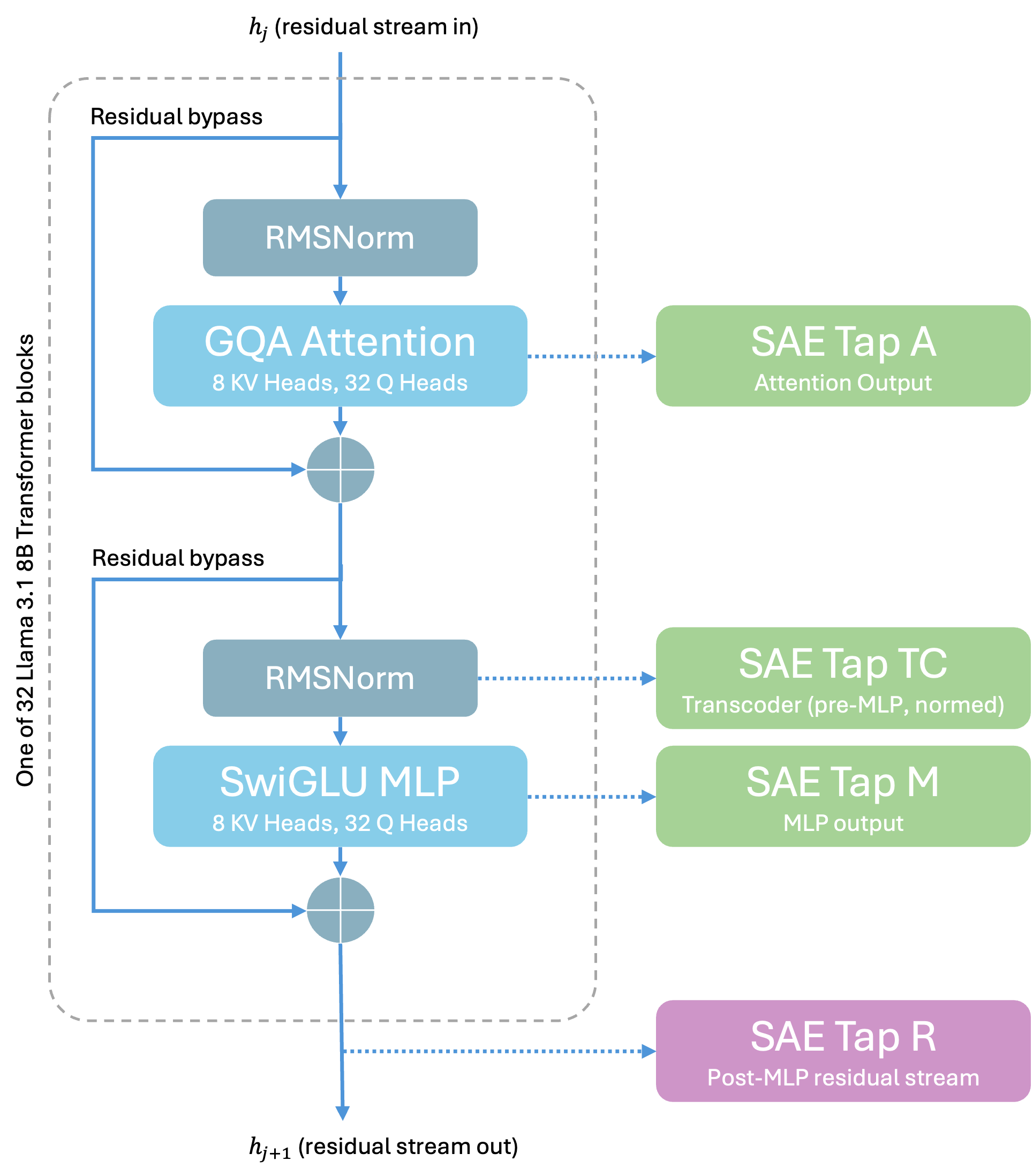}
    \caption{LLama 3.1 8B offers different positions from which representations can be tapped. Of those, R is the most well researched which is used in this work.}
    \label{fig:sae_taps}
\end{figure*}

\subsection{Dataset}
\label{sec:dataset}

Our concept dataset comprises 84 hand-curated concept pairs across four semantic
categories (factual, abstract, overlap, and context-dependent), with the
overlap label deliberately distributed across categories to prevent
category-to-label confounds. Each concept is paired with two or more
cloze-style prompts that elicit the concept in naturalistic context --- for
example, the concept \textit{Photosynthesis} is paired with the prompt
``\textit{The process by which plants convert sunlight into glucose is called
\rule{1cm}{0.4pt}}.'' The model processes each prompt and the last-token hidden
state at the extraction layer is taken as the concept's representation for that
prompt. English pairs constitute 76\% of the dataset, supplemented by
multilingual subsets in German, French, Spanish, and Mandarin Chinese. For
task-level evaluation (\S\ref{sec:task-protocol}), the dataset was expanded to 163 concepts with an
average of 4.7 validated prompts per concept, plus 112
attribute/sense-disambiguation tasks and 107 cross-lingual tasks.

\subsection{Communication Channels}
\label{sec:channels}

Throughout this paper, \textit{sender} is the LLM agent that holds a concept
representation it wishes to communicate; \textit{receiver} is the agent that
must identify or act on that concept; \textit{transmission} is the operation by which
the sender's internal representation is converted into the format required by a
given channel and delivered to the receiver.
Bandwidth figures represent the minimum bits needed to faithfully represent the
transmitted object, not end-to-end network throughput.

\paragraph{Dense latent channel.} Transmits the sender's full residual-stream
hidden state at layer 23 — a 4,096-dimensional bf16 vector (65,536 bits). This
channel incurs zero information loss by construction and serves as the
performance ceiling.

\paragraph{SAE-sparse channel.} First compresses the sender's hidden state
through a pre-trained SAE~\cite{bricken2023monosemanticity} with 131,072 latent
dimensions and $\sim$70 active features per forward pass. Only the indices and
activation magnitudes of active features are transmitted — approximately 2,300
bits, a 28$\times$ compression relative to dense. The SAE weights used are
\texttt{andyrdt/saes-llama-3.1-8b-instruct},
\texttt{resid\_post\_layer\_23\_trainer\_1}. Only the pre-trained SAE decoder
needs to be shared between agents for decoding — not the full sender LLM.

\paragraph{Text channel.} The standard inter-agent baseline. The sender
serialises the concept as its name string (which means it predicts and outputs the "next word" as a string); the receiver processes that string
through the model and extracts the resulting hidden state at layer 23. This
channel requires no model alignment infrastructure.

\begin{figure*}[tb]
    \centering
    \includegraphics[width=.6\linewidth]{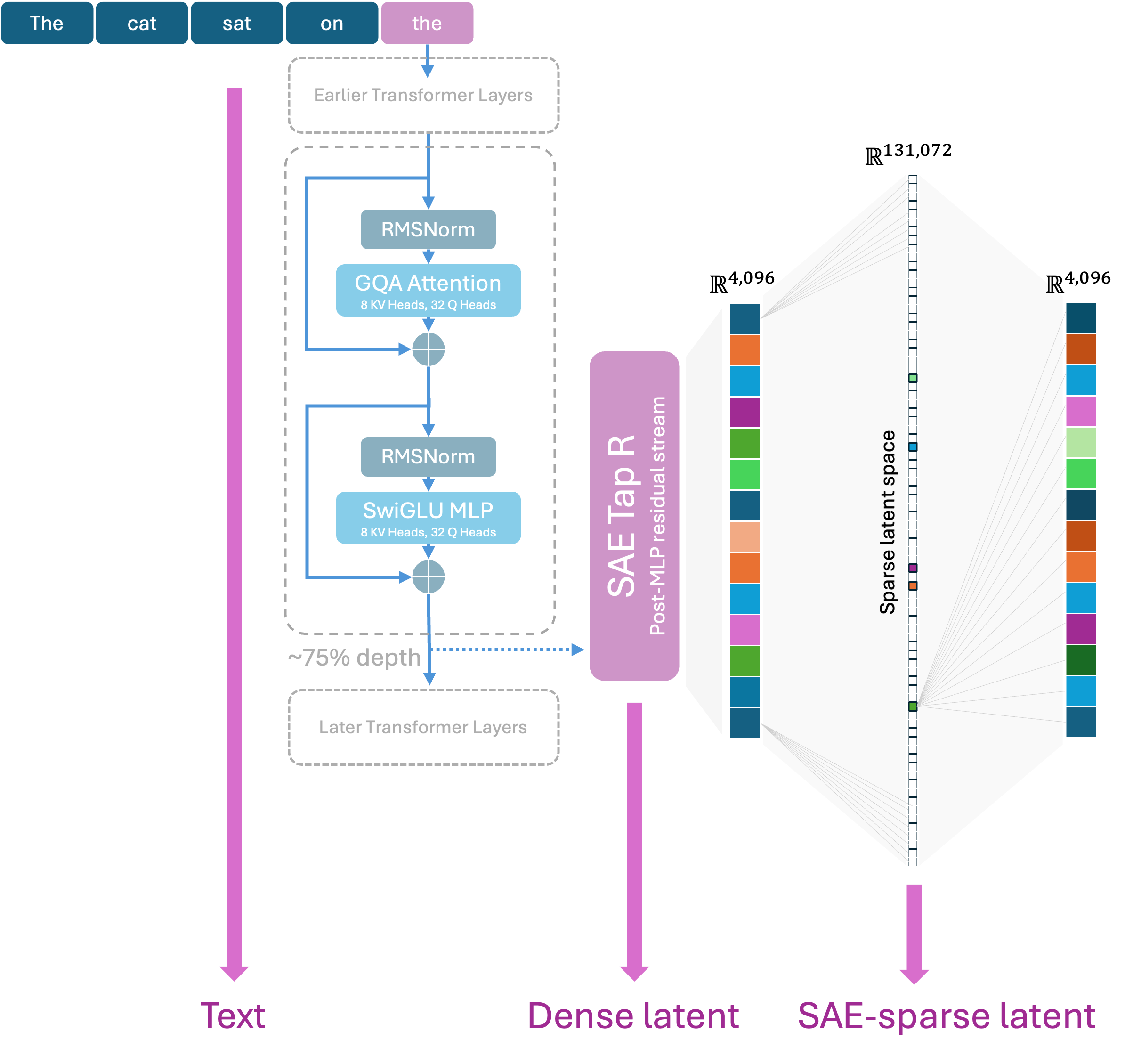}
    \caption{Illustration of communication channels used: Text channel transports concept names or descriptions, the dense latent channel transports the last token internal representation, and the SAE latent channel transports the sparse most activating neurons.}
    \label{fig:comm_channels}
\end{figure*}

The channels differ in what is transmitted, but in all cases the representation
used for evaluation is a 4,096-dimensional hidden-state vector.
For probe-based evaluation (\S\ref{sec:probes}) and task-level evaluation (\S\ref{sec:task-protocol}), this vector is compared geometrically to reference vectors --- no injection into a
running model's forward pass occurs. The SAE-sparse channel differs from the
dense channel only in that the vector passes through an encode--decode
bottleneck: the sparse code (feature indices and magnitudes) is decoded back to
a dense approximation via the SAE decoder matrix. The text channel is the only
channel that re-encodes entirely: the concept string is processed by the
receiver model from scratch, producing a new hidden state that is then used for
comparison.

\subsection{Linear Probes and Channel Fidelity~(C1)}
\label{sec:probes}

In our first claim, we want to elucidate how well the three channels maintain distinctive features in the hidden state representations after transmitting a concept from sender to receiver through the channel. For this, we train logistic regression classifiers on ground-truth sender representations and then tested on versions transmitted through a given channel.

We call these logistic regression classifiers \textit{linear probes}. Given a number of activations $\{h_A^{(i)}\}$ from prompts eliciting concept $A$ and $\{h_B^{(i)}\}$ from concept $B$, the probe learns a linear separating hyperplane. \textit{Probe accuracy} measures how much concept-discriminative information a transmitted representation retains. 

\subsection{SAE Feature Survival Analysis~(C2)}
\label{sec:survival-method}

To characterise text-channel information loss at the Sparse Autoencoder (SAE) feature level, we define a \textit{text round-trip} (details to follow below): a concept is first represented in activation space via a contextual prompt, then, secondly, re-encoded by the same model, but this time reduced to the bare concept name, and the resulting representation is compared with the original.
By then encoding both paths through the SAE, we identify which individual features survive the round-trip and which are replaced.

\textbf{Step 1 — Contextual encoding.} Cloze-style prompts elicit each concept
in rich context (e.g., ``\textit{The process by which plants convert sunlight
into glucose is called \rule{1cm}{0.4pt}.}''). Each prompt is processed through
Llama 3.1 8B-Instruct; the last-token hidden state at layer 23 is extracted and
SAE-encoded. Multiple prompts per concept were represented, and the stable SAE 
features recorded. Over all concepts, we find a per-concept average of $\sim$43 features 
out of the total 131,072 SAE features to be active. Compared to the about 70
active features of one individual forward pass, we infer that the remainder are
not concept-specific.

\textbf{Step 2 — Text-channel encoding.} The bare concept name (e.g.,
\texttt{"Photosynthesis"}) is processed through the same model; the last-token
hidden state at layer 23 is SAE-encoded, again yielding $\sim$43 active
features.

\textbf{Step 3 — Feature set comparison.} For each concept we compute: feature
survival rate, Jaccard overlap, cosine similarity in SAE space, and Pearson
correlation between original and text-channel feature magnitudes.

%TODO: this takes away the suspense... If not needed for the subsequent methods, better save for later.
The fact that both paths produce similar-sized feature sets but nearly disjoint
memberships is the central finding of this analysis.

\subsection{Cross-Architecture Alignment~(C3)}
\label{sec:alignment-method}

Activation spaces are coordinate systems: two models may encode the same concept
as particular directions in their respective 4,096-dimensional spaces, but
nothing forces those directions to coincide. If the two spaces are nonetheless
geometrically compatible — rotated versions of each other — a single rotation
matrix can recover the correspondence from a modest number of paired concept
vectors. These pairs, drawn from concepts whose representations have been
extracted from both models, are called \textit{anchor} concepts; concepts
withheld from the anchor set serve as the held-out test of whether the estimated
rotation generalises beyond the examples it was fitted on.

We test whether two architecturally distinct models encode concepts in
compatible geometric arrangements via Orthogonal Procrustes alignment. For each
anchor concept, both models produce a mean concept vector (averaged over all
available prompts). The Procrustes problem finds the rotation matrix
\begin{equation}
  R^* = \arg\min_{R\,:\,R^\top R = I} \|X R - Y\|_F
\end{equation}
where $X$ and $Y$ are the sender and receiver anchor matrices. The fitted
rotation is applied to held-out concept vectors, and retrieval accuracy is
measured by nearest-neighbour lookup in the receiver's space.

We sweep the anchor count from 50 to 140 and evaluate holdout cosine
similarity, top-1 and top-5 retrieval accuracy, and cross-architecture probe
transfer on held-out concepts. We compare base Mistral (v0.1) and
instruction-tuned Mistral (v0.3) to test whether instruction tuning affects
geometric compatibility.

\subsection{Task-Level Evaluation Protocol}
\label{sec:task-protocol}

While probe accuracy (\S\ref{sec:probes}) measures discriminability in a controlled pairwise setting with access to labelled representations, it does not reflect inference-time conditions. Firstly, a receiver agent neither trains a new classifier for each incoming concept to probe if it is different from others, nor has it access to labelled sender vectors. Therefore, we propose to implement a protocol that is free of such requirements. The following multiple-choice (MC) protocol is designed to mimic the realistic zero-shot setting more closely.

\paragraph{Evaluation mechanism.}
For each task, one concept is transmitted from sender to receiver through a given
channel, producing a query vector $h_\text{query}$ in the receiver's activation
space (Procrustes-mapped for latent channels; re-encoded by the receiver model
from scratch for the text channel). Separately, the receiver pre-caches a
reference vector for each of four candidate concept names by encoding each
candidate as a bare name string through its own model and extracting the
last-token hidden state at layer 24. The receiver then scores all four
candidates by cosine similarity between $h_\text{query}$ and each candidate's
cached reference vector. The highest-scoring candidate is taken as the answer.
No joint forward pass over query and candidates occurs; the evaluation is purely
a nearest-neighbour retrieval in the receiver's activation space.

\paragraph{Channels and baselines.}
All three channels defined in \S\ref{sec:channels} are evaluated, with the text
channel split into two variants to separate the effect of context: (1)~\textbf{dense
latent}: the full sender hidden state, Procrustes-mapped; (2)~\textbf{SAE
latent} (i.e., the SAE-sparse channel): the SAE encode--decode reconstruction,
Procrustes-mapped; (3)~\textbf{text-passage}: the receiver encodes a full
descriptive passage --- e.g.\ ``\textit{The process by which plants convert
sunlight into glucose is called Photosynthesis}'' --- and extracts its own
last-token hidden state; (4)~\textbf{text-name}: the receiver encodes only the
bare concept name, e.g.\ ``\texttt{Photosynthesis}''; and (5)~\textbf{no-comm}:
a baseline in which no signal is transmitted and the receiver scores candidates
using only its cached reference vectors without any query, expected to perform
at chance.

Statistical significance is assessed via one-sided binomial test against
the 25\% chance baseline, which corresponds to four uniformly presented
candidates with no prior over the correct answer, such that a receiver that ignores the
transmitted signal entirely and answers uniformly at random will be correct 25\%
of the time in expectation.

\subsection{Text Augmentation Protocol}
\label{sec:augmentation-method}

The task-level results (\S\ref{sec:task-protocol}) reveal a consistent 3--10\,pp gap in which the text-passage channel outperforms the latent channel. Two hypotheses could explain this: (a)~the Procrustes alignment introduces geometric distortion that corrupts the concept-level signal, in which case adding aligned sender information to the
receiver's text representation should partially close the gap; or (b)~the
receiver's native text processing extracts additional task-relevant signal from
the passage that is simply absent from the sender's latent state, in which case
augmentation provides no benefit. Rather than treating the latent channel as a
replacement for text, the augmentation protocol treats it as a candidate
complement and adjudicates between these hypotheses.

The augmented representation is formed as
\begin{equation}
  h_\text{aug} = h_\text{text} + \alpha \cdot R^* h_\text{sender}
\end{equation}
where $\alpha \in \{0.01, 0.05, 0.1, 0.2, 0.5, 1.0\}$ is a mixing coefficient.
Four variants of $h_\text{sender}$ are tested: the full dense hidden state; the
full SAE encode-decode reconstruction; the top-32 SAE features only; and the
\textit{lost features} — those active in contextual encoding but absent from
bare-name encoding.

% ─────────────────────────────────────────────────────────────────────────────
\section{Results}
\label{sec:results}

\subsection{Single-Model Baselines (Prerequisite Gates)}
\label{sec:baselines}

Before testing the main claims (C1--C3), we verify two prerequisite gates: that
semantic concepts are linearly separable in activation space, and that the SAE
feature space faithfully captures semantic relatedness. Logistic regression
probes trained on layer 15 activations achieve 100\%
binary classification accuracy on all 84 concept pairs in the hand-curated
dataset, confirming linear separability. Activation steering — injecting the
probe's learned normal vector into the residual stream at scale $\alpha=10$ —
produces causal concept flips (e.g., ``Paris'' generative outputs shift to
``London''), confirming causal responsibility of probe directions. Saturation
and semantic degradation occur predictably at $\alpha \geq 50$.

At the SAE feature level, semantically related concept pairs activate
significantly more overlapping features than unrelated pairs:
$\Delta\text{cos} = +0.319$ at layer 23, positive at all seven tested layers
(3, 7, 11, 15, 19, 23, 27). This confirms that the SAE feature space faithfully
encodes semantic relatedness and is a valid measurement instrument for the
channel analysis that follows.

\subsection{Channel Fidelity~(C1)}
\label{sec:c1}

Using the linear probes described in \S\ref{sec:probes}, Table~\ref{tab:channels} summarises the three channels across probe accuracy,
bandwidth, and compression.

\begin{table}[h]
  \centering
  \caption{Channel fidelity across probe accuracy and bandwidth.}
  \label{tab:channels}
  \begin{tabular}{@{}lccc@{}}
    \toprule
    Channel & Probe Acc. & Bandwidth & Compression \\
    \midrule
    Dense      & $100.0 \pm 0.0$\%  & 65,536 b & 1$\times$     \\
    SAE-sparse & $99.4 \pm 3.8$\%   & $\sim$2,300 b & 28$\times$ \\
    Text       & $80.4 \pm 24.4$\%  & $\sim$53 b    & 1,236$\times$ \\
    \bottomrule
  \end{tabular}
\end{table}

The SAE-sparse channel preserves near-perfect concept discriminability (99.4\%)
at 28$\times$ compression. The text channel achieves only 80.4\%, a 19.4\,pp
gap. A generation injection test — SAE-decoded vectors injected into the residual
stream of a neutral prompt at layer 23 — produces the correct concept name in
39.3\% of cases (33/84 injections), a conservative lower bound.

On an information-per-bit basis, text is more efficient: 80.4\% accuracy at 53
bits vs.\ 99.4\% at 2,300 bits. However, text faces an absolute ceiling at
$\sim$80\%, imposed not by bandwidth but by the representational gap examined
next.

\subsection{Feature Survival Analysis~(C2)}
\label{sec:c2}

Text round-tripping (\S\ref{sec:survival-method}) does not attenuate the original features — it replaces them.
Table~\ref{tab:survival} summarises the survival statistics across 165 concepts.

\begin{table}[h]
  \centering
  \caption{SAE feature survival after text round-trip (165 concepts).}
  \label{tab:survival}
  \begin{tabular}{@{}lcc@{}}
    \toprule
    Metric & Mean & Std \\
    \midrule
    Feature survival rate  & 11.7\% & $\pm$6.3\% \\
    Feature loss rate      & 88.3\% & $\pm$6.3\% \\
    Jaccard overlap        &  5.8\% & $\pm$2.6\% \\
    SAE-space cosine       & 0.187  & $\pm$0.116 \\
    Magnitude correlation  & 0.326  & $\pm$0.587 \\
    \bottomrule
  \end{tabular}
\end{table}

\begin{figure*}
    \centering
    \includegraphics[width=.7\linewidth]{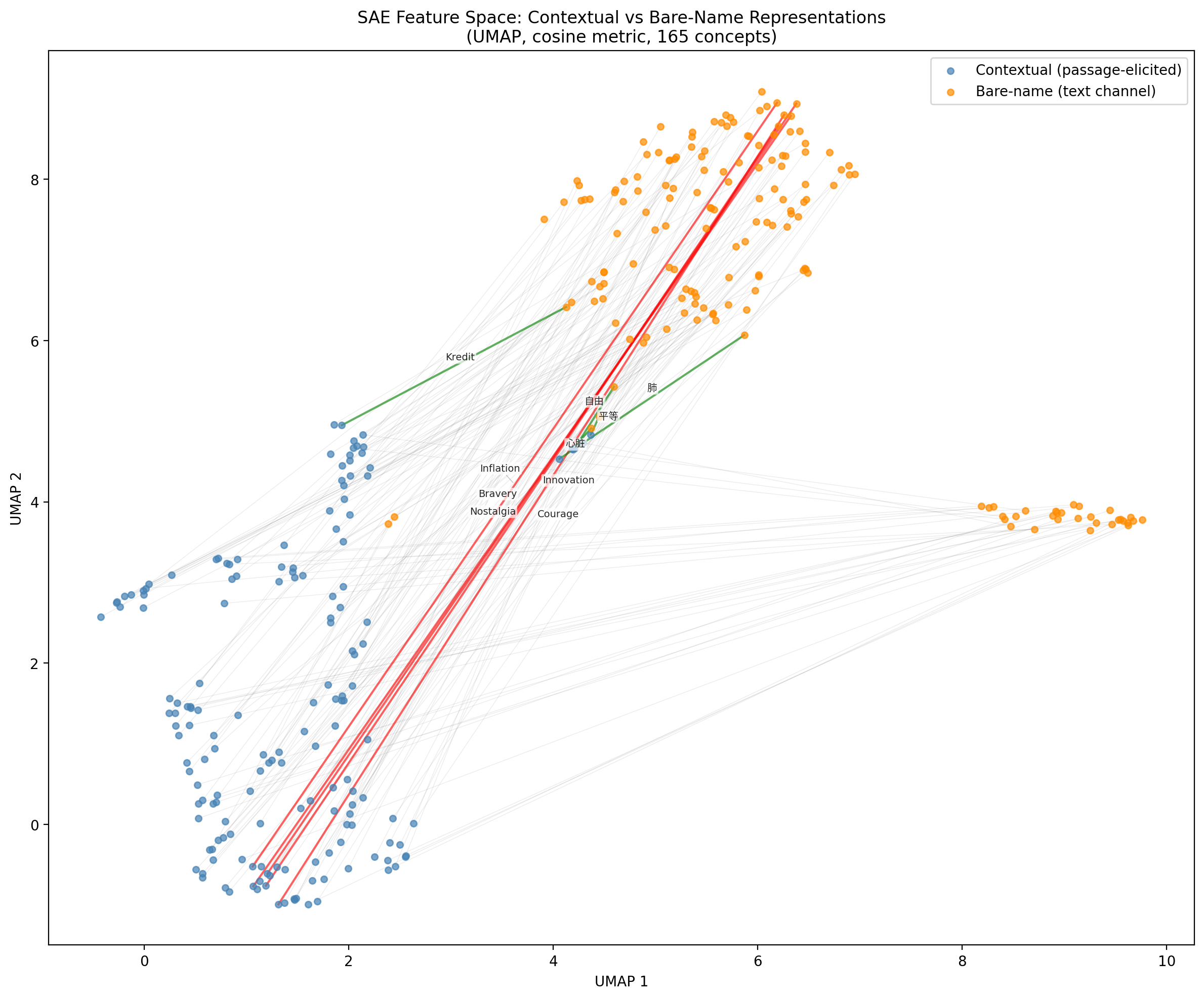}
    \caption{UMAP visualisation of SAE feature vectors; textual annotations indicate most/least displaced concepts.}
    \label{fig:umap}
\end{figure*}

The distinction between attenuation and identity replacement is critical. Under
attenuation, the same features would survive at reduced magnitude — weak signals
falling below an activation threshold. The data rule this out: lost features
have mean activation magnitude 2.36, compared to 2.46 for survived features —
statistically indistinguishable. Features are not disappearing because they are
weak; they are disappearing because the input (the bare concept name in
isolation) simply does not activate them. Text encoding moves the representation
to a different neighbourhood of the 131K-dimensional SAE feature space entirely,
one encoding how the model processes a token label rather than a concept in
context. Figure~\ref{fig:umap} visualises this displacement via UMAP embedding
 of the full SAE feature vectors: contextual and bare-name representations of the same
concept occupy distinct regions of SAE feature space, consistent with wholesale
relocation rather than local perturbation.

The term ``surface form'' as used throughout this paper encompasses three
distinct components of the lost features. First, \textbf{tokenisation
artifacts}: subword chunking differs across models — the same concept string
produces different byte-pair encoding (BPE) splits in the sender and receiver tokenisers, yielding
model-specific embedding patterns that encode \emph{how the input was tokenised}
rather than \emph{what it means}. Second, \textbf{prompt-structural features}:
the template surrounding the concept (system tokens, turn markers, padding)
activates features that reflect the model's processing context rather than
concept semantics. Six SAE features are destroyed 100\% of the time across all
165 concepts, with appearances in 58--111 concepts each — candidate features for
exactly this prompt-structural role. Third, \textbf{positional encoding effects}:
the concept's position within the prompt modulates which features activate,
introducing position-dependent activation patterns that are neither semantic nor
transferable across architectures. Because different models process the same
input differently (different tokenisers, attention patterns, positional
encodings), these three components are inherently model-specific — they would act
as noise if transferred cross-architecture, not as semantic signal.

The survival rate correlates modestly with text-channel probe accuracy (Pearson
$r=0.26$): concepts retaining more features through text encoding tend to be
more discriminable. The modest magnitude indicates that the geometric arrangement
of surviving features matters as much as their count. The augmentation
experiment (\S\ref{sec:augmentation}) provides the decisive evidence on what
these lost features encode: when re-injected into the receiver's representation,
they reduce performance, confirming that the features destroyed by text
serialisation carry surface form — tokenisation artifacts, prompt structure, and
positional encoding effects — rather than concept-discriminative semantics.

\begin{figure}[tb]
    \centering
    \includegraphics[width=1\columnwidth]{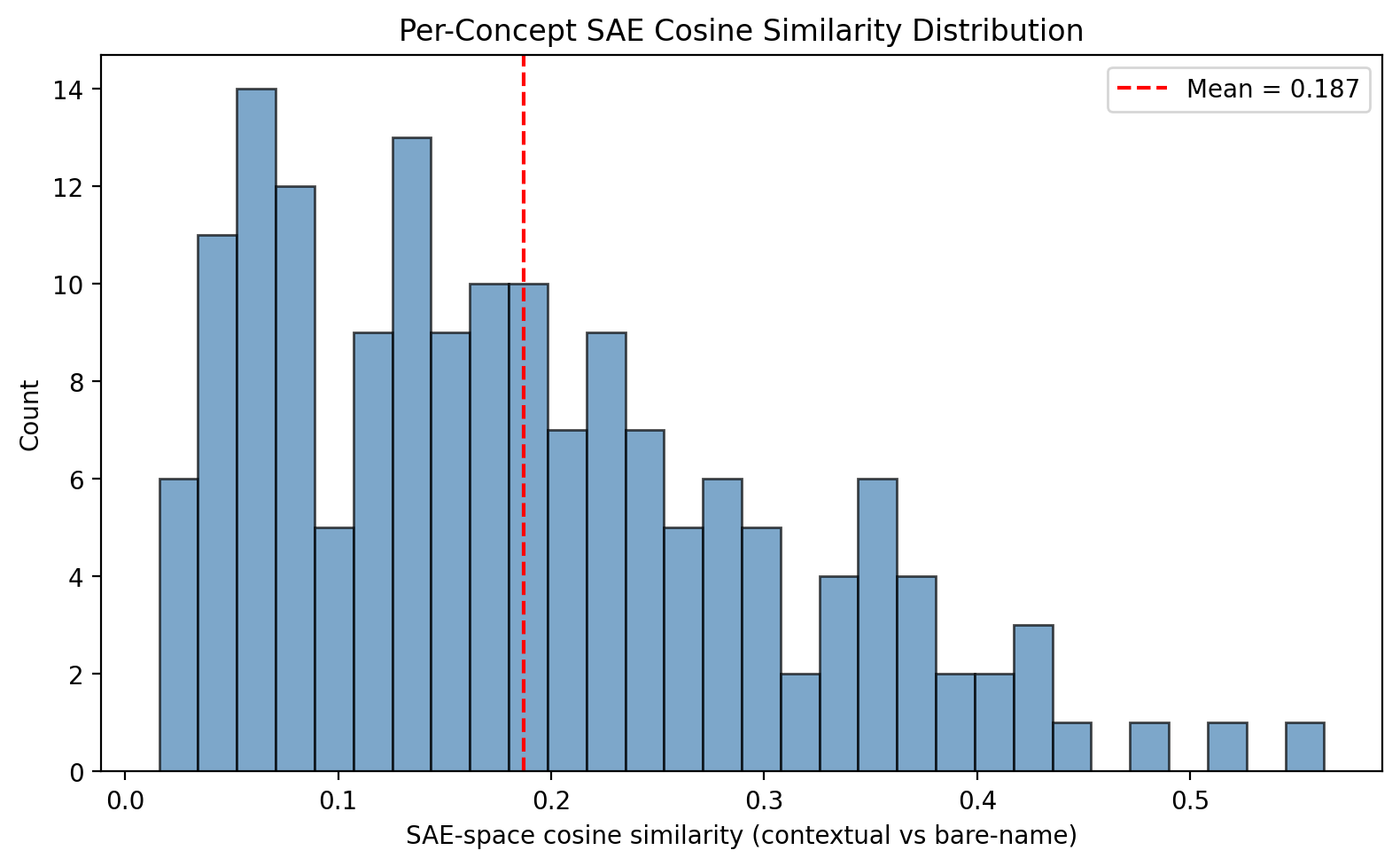}
    \caption{SAE feature cosine similarity distribution through text round-trip.}
    \label{fig:distribution}
\end{figure}

The per-concept SAE cosine similarity distribution (Figure~\ref{fig:distribution})
shows a right-skewed distribution with mean 0.187 and range [$\sim$0.02,
$\sim$0.55], confirming that identity replacement is the norm rather than the
exception, though a small tail of concepts retains modest feature overlap
through the text round-trip.

\subsection{Cross-Architecture Alignment~(C3)}
\label{sec:c3}

Following the Procrustes methodology described in \S\ref{sec:alignment-method}, unaligned cosine
similarity between Llama and Mistral concept vectors is
$-0.002$, indistinguishable from random. After Procrustes alignment, dense top-1
retrieval scales strongly with anchor count (Table~\ref{tab:procrustes}).

\begin{table}[h]
  \centering
  \caption{Procrustes alignment retrieval vs.\ anchor count (Llama $\to$
    Mistral base v0.1).}
  \label{tab:procrustes}
  \begin{tabular}{@{}ccccc@{}}
    \toprule
    $n$ anchors & Holdout $n$ & Cos & Top-1 & Top-5 \\
    \midrule
     50  & 115 & 0.416 & 43.5\% & 81.7\%  \\
     80  &  85 & 0.491 & 75.3\% & 94.1\%  \\
    100  &  65 & 0.517 & 84.6\% & 96.9\%  \\
    120  &  45 & 0.529 & 82.2\% & 97.8\%  \\
    \textbf{140} & \textbf{25} & \textbf{0.530} & \textbf{92.0\%} & \textbf{100\%} \\
    \bottomrule
  \end{tabular}
\end{table}

At 100 anchors, retrieval reaches 84.6\% top-1 — 97$\times$ above the 0.87\%
random baseline. The SAE-reconstructed channel scales in parallel (72.3\% $\to$
80.0\% top-1 at 100 $\to$ 140 anchors), confirming that cross-architecture
geometric structure survives the sparse bottleneck.

Base Mistral (v0.1) aligns slightly better than instruction-tuned Mistral (v0.3)
at every anchor count (92.0\% vs.\ 88.0\% at 140 anchors). This is consistent
with the Platonic Representation Hypothesis: alignment quality reflects
pretraining convergence on shared data statistics, not instruction-tuning
objectives. RLHF does not improve cross-architecture geometric compatibility.

Cross-architecture probe transfer rises from 50.0\% (chance) to 60.5\% when
evaluated with the expanded concept dataset (4.7 prompts per concept, averaged).
The dataset was the bottleneck: with only 2 prompts per concept, 4,096-dimensional
decision boundaries are fundamentally underdetermined.

As a diagnostic, we compared Procrustes with Canonical Correlation Analysis
(CCA), which identifies maximally correlated linear subspaces rather than a
single rotation. CCA achieves 100\% top-1 retrieval at 80 anchors in a
79-dimensional shared subspace, compared to 75.3\% for Procrustes in the full
4,096-dimensional space. This confirms that the cross-architecture semantic
correspondence is concentrated in a low-rank subspace and that the Procrustes
gap reflects alignment method limitations rather than fundamental geometric
incompatibility. A 2-layer MLP (non-linear alignment) fails completely at all
anchor counts due to catastrophic overfitting on $\leq$140 training pairs in
4,096 dimensions; testing whether a non-linear component exists in the
cross-architecture transformation requires substantially larger anchor sets.

\subsection{Task-Level Evaluation}
\label{sec:task-results}

\paragraph{Concept identification (165 tasks).}
Dense latent: 58.2\%. Text-passage: 65.5\% (chance: 25\%). The 7.3\,pp gap is consistent
across concept categories. Using prompt-averaged sender vectors, the dense
latent channel rises to 62.4\%, narrowing the gap to 3.1\,pp.

Within-concept representation variance provides a partial explanation for the
latent channel's disadvantage. Passage-elicited sender representations show
substantial prompt dependence: mean pairwise cosine similarity between different
prompts for the same concept is 0.60 (Llama) to 0.53 (Mistral), compared to 1.0
for the deterministic bare-name text encoding. This variance is
concept-intrinsic — the cross-model correlation of per-concept variance is
$r = 0.92$ — with abstract and polysemous concepts (e.g., ``Speed'',
``Compassion'') showing the highest variability and concrete concepts (e.g.,
``Heart'', ``Chlorophyll'') the lowest. Sender variance correlates weakly with
task failure across all channels ($r \approx -0.16$ for latent, $-0.11$ for
text), indicating that the latent channel is modestly more sensitive to
prompt-level noise. This is consistent with the prompt-averaging result:
averaging over multiple sender prompts narrows the latent--text gap from
7.3\,pp to 3.1\,pp by cancelling prompt-specific activation components.

\paragraph{Attribute and sense disambiguation (112 tasks).}
Table~\ref{tab:tasks} reports results across channels and task sub-types. The
gap widens to $\sim$10\,pp on richer tasks. SAE latent matches dense latent to
within 1\,pp — the sparse bottleneck does not itself degrade task performance;
the cost is entirely attributable to Procrustes alignment quality.

\begin{table}[h]
  \centering
  \caption{Task-level evaluation on 112 attribute and sense disambiguation
    tasks.}
  \label{tab:tasks}
  \begin{tabular}{@{}lccc@{}}
    \toprule
    Channel & Overall & Attr.\ ($n$=52) & Sense ($n$=60) \\
    \midrule
    Dense latent  & 67.0\% & 65.4\% & 68.3\% \\
    SAE latent    & 67.9\% & 65.4\% & 70.0\% \\
    Text-passage  & 76.8\% & 76.9\% & 76.7\% \\
    Text-name     & 39.3\% & 50.0\% & 30.0\% \\
    No-comm       & 25.9\% & 28.8\% & 23.3\% \\
    \bottomrule
  \end{tabular}
\end{table}

\paragraph{Cross-lingual concept identification (74 tasks, 5 languages).}
Dense latent: 64.9\%. Text-passage: 63.5\% (chance: 25\%). This is the only tested regime in
which the latent channel matches text (+1.4\,pp). The Procrustes geometry
preserves language-agnostic semantic structure: the sender's Llama representation
of ``Herz'' (German) maps to a region of Mistral's activation space closer to
the correct English concept than the raw cross-lingual text-passage baseline.

\paragraph{Cross-lingual sense disambiguation (33 tasks).}
Dense latent: 51.5\%, barely above the 25\% chance baseline.
Text-passage: 75.8\% ($-$24.2\,pp). The latent channel
fails at sense-level disambiguation across languages. The Procrustes mapping was
fitted on English concept-level anchors and does not preserve fine-grained
polysemy distinctions reliably. Text-name collapses to 27.3\% on these tasks —
the bare ambiguous token (e.g., ``Avocat'') carries no sense — confirming that
the task requires contextual information and the sense failure of the latent
channel is a genuine deficit, not a task design artifact.

\subsection{Text Augmentation}
\label{sec:augmentation}

Augmenting the receiver's text-passage representation with Procrustes-mapped
sender features provides no reliable benefit. On English tasks, all four
augmentation variants match the 76.8\% text-passage baseline at low $\alpha$
and degrade monotonically above $\alpha=0.2$. On cross-lingual tasks, text+dense
and text+sae\_all show a marginal +0.9\,pp at $\alpha=1.0$ (one task flip out of
107 — not a reliable effect at this magnitude). The lost-feature variant
actively reduces performance: from 66.4\% at $\alpha=0.01$ to 57.0\% at
$\alpha=1.0$. This is the decisive result: the features destroyed by text
serialisation, when re-introduced, act as noise. They are not the carriers of
suppressed semantic signal; they are the carriers of surface form that is
incompatible with the receiver's downstream processing.

\subsection{Summary of Gates and Claims}
\label{sec:summary}

Table~\ref{tab:summary} summarises the outcome of every gate and claim.

\begin{table}[h]
  \centering
  \caption{Summary of experimental gates and claims.}
  \label{tab:summary}
  \begin{tabular}{@{}llr@{}}
    \toprule
    Claim / Gate & Result & Key Number \\
    \midrule
    Separability (\S\ref{sec:baselines})        & Confirmed  & 100\% probe acc. \\
    SAE overlap (\S\ref{sec:baselines})          & Confirmed  & $\Delta$cos = +0.319 \\
    Feature loss (\S\ref{sec:c2})        & Characterised & 88.3\% destroyed \\
    Sparse channel (\S\ref{sec:c1})       & Confirmed  & 99.4\% vs 80.4\% \\
    Cross-arch (\S\ref{sec:c3})           & Confirmed  & 92\% top-1 \\
    Cross-arch probe transfer   & Passed     & 60.5\% \\
    Cross-lingual concepts      & Passed     & +1.4\,pp over text \\
    Cross-lingual sense         & Failed     & $-$24.2\,pp \\
    Text augmentation           & Null       & 0\,pp benefit \\
    \bottomrule
  \end{tabular}
\end{table}

% ─────────────────────────────────────────────────────────────────────────────
\section{Discussion}
\label{sec:discussion}

\subsection{What We Found}

\paragraph{Representational convergence is real.}
Llama 3.1 8B-Instruct and Mistral 7B develop activation geometries alignable by
a simple orthogonal rotation, yielding 92\% top-1 retrieval at 140 anchors
against a 0.87\% chance baseline. This is strong empirical support for the
Platonic Representation Hypothesis: architectures with different tokenisers,
attention structures, and training lineages converge toward compatible semantic
geometry because they model the same distributional properties of language.

\paragraph{SAE features quantify what text loses — and what is lost is not the
semantics.}
Text serialisation performs catastrophic feature replacement (88\% of contextual
features replaced, Jaccard 5.8\%, SAE cosine 0.187). But the augmentation
experiment resolves the consequential question: the destroyed features are not
the ones encoding concept-discriminative information. When re-injected, they
degrade performance. The loss is structural — text cannot carry prompt-contextual
and relational representations — but for concept-identification tasks, this
structural loss does not cross the threshold of task relevance.

\paragraph{The Procrustes tax is real but tractable.}
The consistent 3--10\,pp gap between the latent and text channels reflects
current alignment quality, not a fundamental geometric ceiling. Averaged sender
vectors close much of the gap; better alignment methods — CCA, learned linear
projections, or nonlinear approaches — would likely reduce it further.

\subsection{The Honest Negative}

On every task type tested, the text channel matches or exceeds the latent
channel. The cross-lingual parity result (+1.4\,pp) is the closest the latent
channel gets to outperforming text. Text augmentation confirms this from the
opposite direction: adding latent information to an already-text-capable receiver
is useless at best.

This is informative precisely because the experimental infrastructure is sound.
The null result cannot be attributed to poor alignment (92\% retrieval), poor
channel fidelity (99.4\% SAE accuracy), or insufficient data (probe transfer
rises from 50\% to 60.5\% with more prompts per concept). The limitation is that
all tested tasks are fully text-expressible. A receiver with access to a rich
text passage already holds all concept-discriminating signal that the current
task format requires.

\subsection{When Would Latent Channels Win?}

Three conditions could in principle yield a practical latent advantage, none of
which are met by the current experiments:

\begin{enumerate}
  \item \textbf{Tasks requiring information that text cannot express.} If a
    model's latent representations encode relational or structural knowledge that
    resists faithful serialisation to a token sequence, the text channel would
    face a performance ceiling well below 90--100\%. No current task type probes
    this.
  \item \textbf{Cross-modal communication with a pure vision model.} A model
    trained exclusively on images cannot re-encode concepts from text
    descriptions — the ``text channel'' path simply does not exist. Any geometric
    alignment must be emergent, not engineered.
  \item \textbf{Bandwidth-constrained or interpretability-constrained settings.}
    The SAE channel's 28$\times$ compression over dense — at 99.4\% fidelity and
    with named, interpretable feature activations — has practical value
    independent of whether it exceeds text accuracy.
\end{enumerate}

\subsection{Limitations}

Both sender and receiver are 8B-parameter models; larger models may develop
richer latent structures in which contextual representations encode a greater
proportion of task-relevant information. There are only very few pretrained SAEs available for such larger models, so that for this initial investigation we chose to restrict ourselves to small models.
% TODO: might check transfer between large to small to validate that the small model can't differentiate a larger model's latents?

Also, orthogonal Procrustes is not the
optimal linear alignment method for the cross-architecture setting: CCA
retrieval experiments show that an anisotropic linear alignment achieves
near-perfect concept retrieval (\S\ref{sec:c3}), confirming that the
cross-architecture correspondence is concentrated in a low-rank subspace.
However, our augmentation results (\S\ref{sec:augmentation}) suggest that even perfect alignment
would not cause the latent channel to exceed text on concept-identification
tasks: the information text loses is surface form, not task-relevant semantics.
Improved alignment would likely close the 3--10\,pp gap without creating a
latent advantage on current task types. 

Lastly, the prompt-variance analysis is based on only 2--4 prompts per concept; finer-grained variance estimates would require 10+ prompts per concept. The dataset also only covers 165 general-knowledge English concepts; domain-specific vocabularies may have different feature survival profiles. Finally, all task types are designed with text-expressible answers \textemdash\ which is precisely the condition under which the text channel performs best.

\paragraph{Structural injection barriers in instruction-tuned models.}
Pilot experiments with Gemma~3 27B-IT reveal that latent injection into
instruction-tuned models with structured prompt templates faces architectural
obstacles beyond geometric alignment. The BOS token position acts as an
attention sink, accumulating a hidden-state norm 35$\times$ larger than regular
positions by layer~20. Injecting sender vectors at this position produces
$\leq$3.3\% accuracy regardless of blending coefficient, because the sender
signal constitutes $<$7\% of the blended vector's energy even at $\alpha=0.70$.
Norm-rescaling recovers partial signal (10\%) but remains far below the only
viable alternative: last-token injection at layer~20, which achieves 20\% on an
85-concept evaluation \textemdash\ above the no-communication baseline (McNemar $p < 0.001$)
but not significantly above chance. More fundamentally, embedding-level prepend
of source tokens before the chat template markers fails completely (3.3\%),
despite the same tokens achieving 96.7\% when placed \emph{within} the template.
Instruction-tuning couples semantic extraction tightly to template structure;
tokens outside the expected format are treated as positional noise. Any model
with structured prompt templates \textemdash\ which includes all current
instruction-tuned and chat-finetuned architectures \textemdash\ will have analogous
structural anchor positions where direct injection fails. Practical deployment
of latent injection in multi-agent systems must account for these template-level
constraints.

% ─────────────────────────────────────────────────────────────────────────────
\section{Future Work}
\label{sec:future}

\subsection{Deeper-Than-Language Knowledge}

The central open question this paper cannot answer is whether a language model
encodes knowledge in its latent space that it cannot faithfully transmit through
text. We conducted preliminary Chain-of-Thought (CoT) experiments to probe
this. On BIG-Bench Hard logical reasoning tasks, we compared task accuracy when
a sender communicates via its pre-articulation hidden state versus the full
text CoT trace.

On a 27B-parameter model (Gemma~3), CoT improves accuracy from 70.7\% (direct)
to 87.2\% (with reasoning), identifying 157 ``reverse-gap'' tasks where CoT
succeeds but direct inference fails. Injecting the sender's pre-answer hidden
state into the receiver recovers 33.1\% accuracy on these tasks --- well above
the 0.6\% no-communication baseline, confirming that latent states carry
reasoning-relevant signal. However, the text CoT channel achieves 96.2\% on the
same tasks, a 63\,pp advantage over latent injection.

These results illustrate both the promise and the difficulty: latent states do
encode reasoning information that is absent from the direct answer, but the
text serialisation of that reasoning is far more effective than the latent
channel at conveying it. Demonstrating a latent advantage on reasoning tasks
would require either tasks where CoT serialisation itself loses information, or
substantially improved injection methods. Both directions require larger models
and additional compute beyond our current experimental setup.

\subsection{Cross-Modal Medical Alignment}

A more immediate and equally rigorous test of the Platonic hypothesis is
cross-modal alignment: does a pure vision model and a pure language model
develop compatible representations of the same medical concepts?

The critical constraint is that the vision model must be trained exclusively on
image supervision — no text encoder, no contrastive text-image loss, no language
pretraining. Shared geometry in such a system cannot be attributed to joint
training; it would be purely emergent. We identify MedicalMultitaskModeling (M3
/ UMedPT; \cite{schaefer2024medfoundation}) as the candidate: a foundational
medical imaging framework trained via supervised multi-task learning on
segmentation, classification, and detection across diverse biomedical imaging
modalities, with no text objective.

The alignment protocol follows the cross-architecture Procrustes method
(\S\ref{sec:alignment-method}), adapted for the dimensionality mismatch
between the vision encoder's feature space and the language model's residual
stream. Orthogonal Procrustes requires equal-dimensional spaces; CCA or learned
linear projections will be required. Anchor concepts must be represented by both
a canonical imaging example and a text description. A positive retrieval result
— given a pathology image, retrieve the correct text concept — would be strong
evidence for cross-modal representational convergence: two models that have never
shared a modality developing aligned semantic geometries because they model the
same underlying domain structure.

% ─────────────────────────────────────────────────────────────────────────────
\section{Conclusion}
\label{sec:conclusion}

We have constructed and evaluated the first systematic measurement framework for
latent versus text communication between large language model agents. Three
communication channels — dense latent, SAE-sparse, and text — were compared
across representation-level fidelity, cross-architecture alignment quality, and
task-level concept discrimination.

Representational convergence between Llama 3.1 8B-Instruct and Mistral 7B is
unambiguously confirmed: Procrustes alignment achieves 92\% top-1 concept
retrieval, far above the 0.87\% chance baseline. SAE feature analysis reveals
that text serialisation performs catastrophic feature replacement (88\% of
features destroyed), but augmentation experiments establish that these lost
features carry surface-form information rather than concept-discriminative
semantics. On all tested task types, text communication is sufficient — the
latent channel matches but does not exceed it.

The open question is not whether latent communication works, but when it is
necessary. The current experiments establish what text-expressible tasks can
reveal: convergence is real, the loss is structural but task-irrelevant for
current benchmarks, and the Procrustes tax is tractable. The two frontiers that
define the next phase of inquiry are tasks probing knowledge genuinely beyond
textual expressibility, and cross-modal alignment with pure vision models — the
latter being both practically motivated as a clinical tool and a rigorous test
of representational convergence in its strongest cross-modal form.

% ─────────────────────────────────────────────────────────────────────────────

\end{document}